\documentclass{article}

\usepackage[final]{neurips_2024}

\usepackage[utf8]{inputenc} 
\usepackage[T1]{fontenc}    
\usepackage{hyperref}       
\usepackage{url}            
\usepackage{booktabs}       
\usepackage{amsfonts}       
\usepackage{nicefrac}       
\usepackage{microtype}      
\usepackage{xcolor}         
\usepackage{graphicx}
\usepackage{array}
\usepackage{longtable}
\usepackage{subcaption}

\title{MIMIC: \underline{M}ultimodal \underline{I}slamophobic \underline{M}eme \underline{I}dentification and \underline{C}lassification}

\author{%
S M Jishanul Islam$^{1}$\thanks{Equal Contribution} \quad Sahid Hossain Mustakim$^{1*}$ \quad Sadia Ahmmed$^{1*}$ \\ \quad \textbf{Md. Faiyaz Abdullah Sayeedi}$^{1*}$ \quad \textbf{Swapnil Khandoker}$^{2*}$
\textbf{Syed Tasdid Azam Dhrubo}$^{3*}$ \\ \quad \textbf{Nahid Hossain}$^1$\\
$^1$United International University \quad $^2$Johannes Kepler Universität, Linz \quad $^3$University of Alberta\\
\texttt{\{sislam201024,smustakim201274,sahmmed201146,msayeedi212049\}@bscse.uiu.ac.bd}\\
\texttt{k12215556@students.jku.at}\\
\texttt{syedtasd@ualberta.ca}\\
\texttt{nahid@cse.uiu.ac.bd}
}

\begin{document}

\maketitle

\vspace{-0.3cm}

\begin{abstract}
Anti-Muslim hate speech has emerged within memes, characterized by context-dependent and rhetorical messages using text and images that seemingly mimic humor but convey Islamophobic sentiments. This work presents a novel dataset and proposes a classifier based on the Vision-and-Language Transformer (ViLT) specifically tailored to identify anti-Muslim hate within memes by integrating both visual and textual representations. Our model leverages joint modal embeddings between meme images and incorporated text to capture nuanced Islamophobic narratives that are unique to meme culture, providing both high detection accuracy and interoperability.

\end{abstract}

\section{Introduction}

The widespread use of social media has transformed memes into a popular form of digital communication. While memes are often created for humor, they can serve as powerful vehicles to spread hate speech and reinforcing harmful stereotypes. The field of hate speech on social media platforms has become increasingly sophisticated through the use of memes—multimodal content that combines images and text to spread harmful narratives. While progress has been made in detecting general hate speech (\cite{subramanian2023survey}), the specific challenge of identifying and countering anti-Muslim hate memes remains largely unaddressed. Recent advances in multimodal learning have demonstrated promising results in meme classification tasks (\cite{bikram2024memeclip}). However, these developments are hindered by a critical limitation: the absence of datasets focusing on anti-Muslim hate memes. Existing hate speech datasets (\cite{hermida2023detecting}) either focus solely on text-based content or address broader categories of direct hate speech, failing to capture the covert form of hate with cultural nuances specific to anti-Muslim prejudice expressed through memes. 

To address this gap, we present a novel dataset of anti-Muslim hate memes collected from various online platforms.  Our research reveals distinct patterns in how anti-Muslim sentiment is propagated through memes, highlighting the importance of considering both cultural context and multimodal elements in hate speech detection systems. These insights not only advance our understanding of online Islamophobia but also provide practical implications for content moderation strategies. The code and dataset are open-sourced\footnote{Code and Data: \url{https://github.com/faiyazabdullah/MIMIC}}.

\section{Related Works}

\label{literature_review}

Hate speech detection on social media has become a critical area of research, particularly with the rise of multimodal content that combines text and imagery to convey offensive or discriminatory messages (\cite{arya2024multimodal}). While early studies on hate speech detection relied primarily on text-based datasets, advances in deep learning have allowed researchers to expand beyond text, employing multimodal approaches that incorporate visual elements and language models to improve accuracy and contextual understanding (\cite{guo2023investigation}). In response to the limitations of text-only approaches, recent research has focused on multimodal hate speech detection, particularly in the context of memes (\cite{gandhi2024hate}). Memes present a unique challenge, as they often blend image, text, and context-dependent humor to convey subtle or overt hate messages. Visual language models (VLMs) and transformer-based architectures such as VisualBERT (\cite{li2019visualbert}), ViLBERT (\cite{lu2019vilbert}), and CLIP (\cite{radford2021learning}) have shown promise in addressing these challenges. MemeCLIP (\cite{shah2024memeclip}) was designed for multimodal hate detection, demonstrating that integrating visual and textual representations improves model performance in identifying hate memes. Although similar models achieve high accuracy in general hate meme classification, they lack specificity for certain types of hate, particularly Islamophobic content. 

Recent developments in visual language models (ViLMs) and optical character recognition (OCR) techniques have enhanced multimodal hate speech detection capabilities. (\cite{kim2021vilt}) introduced ViLT, which is very effective for visual question answering and meme analysis tasks. Fine-grained OCR model by (\cite{pettersson2024multimodal}), has improved text recognition in complex, low-quality images, enabling more accurate text extraction in memes with varying font styles, languages, and image quality.

\section{Dataset}
\label{dataset}


Our dataset consists of 953 memes gathered from Reddit, X, 9GAG, and Google Images, capturing diverse examples of potential anti-Muslim content. These memes were carefully curated to represent a range of content with potential anti-Muslim sentiment. We only take the samples which have text incorporated in the images, as we formally define them as "memes". To label the dataset, the annotators comprised of researchers with experience in hate speech detection, and conducted a manual review of each meme to classify it as hateful or non-hateful towards Muslims. Annotation used binary classification (0: non-hateful, 1: hateful), with 545 non-hateful and 408 hateful labels. To reduce bias, we established predefined rubrics based on language, symbols, and context, helping annotators make consistent decisions. Disagreements were addressed through discussions, with a consensus threshold requiring at least 80\% annotator agreement to confirm a label. The distribution of labels and statistics are shown in Figure~\ref{fig:dataset_analysis}, with detailed dataset information presented in Table~\ref{tab:image_table}.

\section{Methodology}
\label{methodology}

\begin{figure}[t!]
    \centering
    \includegraphics[width=\textwidth]{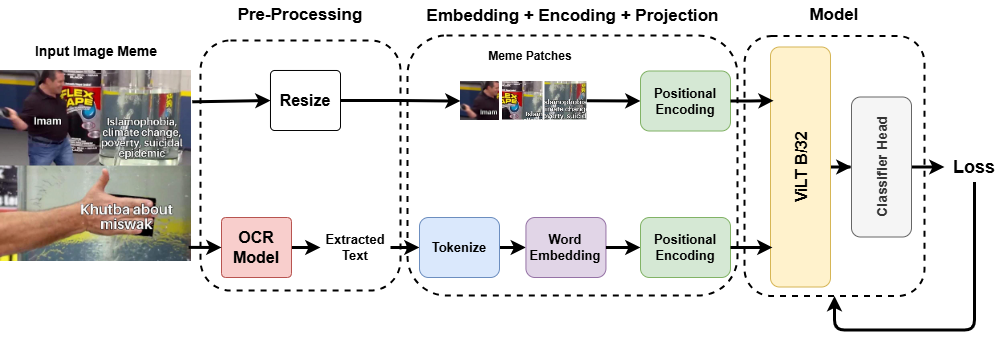}
    \caption{The end-to-end pipeline of our methodology}
    \label{fig:methodology}
\end{figure}

This section presents our methodology for classifying memes into hateful and non-hateful categories. The end-to-end pipeline of our methodology is shown in Figure~\ref{fig:methodology}.

\subsection{Data Pre-Processing}

Initially, we extract the text in the memes using an optical character recognition (OCR) model by (\cite{wei2024general}), which extracts fine-grained OCR from images. To avoid dimension errors, we ensure that the texts are of the same length during batch-wise training by padding the texts to the maximum length of the extracted text from a meme, which is 40. We resize the images to $252 \times 252$ to ensure that the pixel values and attention mask generated by the models' preprocessor are consistent for every image data. Additionally, we applied the random rotation data augmentation technique to cover up for the small dataset size. We use this specifically as it does not distort the image or reverse the text in the memes. We record and compare its performances in Table \ref{tab:augvsnonaug}.

\subsection{Visual Language Model}

To learn the representations between the image meme and the OCR-extracted caption, we utilize the Vision-and-Language Transformer (ViLT) base model proposed in (\cite{kim2021vilt}), primarily used for visual-question answering. We use ViLT because it is a transformer-based architecture (\cite{vaswani2017attention}) designed to handle vision-language tasks by simplifying the representation by directly integrating image and text modalities without relying on convolutional neural networks (CNNs) or region-based detectors. It directly projects the raw patches of the meme image and a linear embedding for the OCR-extracted meme text to prepare them for modality interaction. This means there is no preliminary step in our method to extract the image embeddings using a CNN backbone (\cite{huang2020pixel, he2016deep, xie2017aggregated}). Avoiding this visual backbone reduces the computational overhead of our method. Positional encodings are added to the text and image embeddings. Next, the image and text embeddings are concatenated along the sequence dimension to form a unified input representation for the transformer, where self-attention mechanisms capture relationships within the meme image and the text in it. We record the performance of the ViLT model compared to alternatives in Section \ref{experiments}.

\subsection{Classifier Head}

The features learned from ViLT are pooled and passed to a classifier head with sequential multi-layer perceptions to refine and map representations to the output space. It starts with layer normalization, dropout (0.3), and a fully connected layer projecting to 768 dimensions, matching the ViLT output. Another normalization layer, followed by ReLU activation and dropout, introduces non-linearity and regularization. Finally, a fully connected layer outputs predictions, with a sigmoid activation bounding the output between 0 and 1 for hateful/non-hateful classification.

\section{Experiments}
\label{experiments}
\vspace{-0.3cm}
This section presents the experiments conducted to test the performance of our model in hateful Islamic meme classification. We describe the experimental setup and record the median performance of our model with two dataset split settings. 

\subsection{Experimental Setup}
\vspace{-0.3cm}
The experimental setup outlines the hyperparameter and device configurations, and the evaluation metrics used to validate the effectiveness of our proposed approach. The experiments are carried out in a Kaggle environment with an NVIDIA P100 GPU with 16 GB memory. The model is trained and evaluated with two different independent techniques: splitting the dataset into train:validation:test set, and conducting a k-fold cross-validation. The training was done for 10 epochs using the Adam (\cite{kingma2014adam}) optimizer with a learning rate of $1e^{-4}$with a batch size of 16 for training samples, and 2 for the validation and test samples. The loss function used is the binary cross-entropy loss. A batch size of 16 was selected to balance computational efficiency with model performance during training. The issue of overfitting was mitigated by implementing early stopping and regularization techniques during training. The execution time averaged 3 hours, underscoring the computational demands of multimodal analysis.

\subsection{Evaluation Metrics}
\vspace{-0.3cm}
The model's performance is evaluated using the $F_{1}$ score. Moreover, the macro and micro average scores are also recorded. We select the $F_{1}$ score due to the class imbalance present in the dataset. The model's performance is further assessed using $Precision$, which measures the proportion of correctly identified positive instances among all instances predicted as positive. $Recall$ indicates the proportion of correctly identified positive instances out of all actual positive instances in the dataset.

\subsection{Results}
\vspace{-0.3cm}
Table \ref{tab:modelanalysis} shows that the ViLT performs better than alternative visual-language models such as CLIP (\cite{radford2021learning}) and VisualBERT (\cite{li2019visualbert}). The results shown in Table~\ref{tab:traintesttable} summarize the model's performance on the test set, following a train-validation-test dataset split. The model achieves a median loss of 0.621. The $precision$ is relatively high, with a median score of 0.872, indicating that 87.2\% of the positive predictions made by the model are correct. However, the $recall$ is recorded to have a median value of 0.617, exhibiting room for improvement. This suggests that the model correctly identifies 61.7\% of the actual positive instances. The median weighted $F_{1}$-weighted score is 0.581, reflecting a moderate overall performance. While this score indicates reasonable performance, it also underscores the model's challenges in achieving perfect generalization. The training and validation curves in Figures \ref{fig:f1weighted} and \ref{fig:losscurves} further illustrate signs of overfitting, as the model exhibits significantly better performance on the training data compared to the test set. This can be due to several issues, such as insufficient data and noise within the dataset. 

To address potential overfitting and assess the model's generalization ability, we evaluate the model using the K-fold cross-validation technique. We begin by splitting the dataset into a 90:10 ratio, where 90\% is divided into K-folds for training and evaluation, while the remaining 10\% serves as a holdout set to test the model on unseen data. The results are shown in Table~\ref{tab:kfoldtable}. Overall, K-fold cross-validation outperforms the traditional train-validation-test split approach. Specifically, the model achieves an $F_{1}$-weighted score of 0.716 for K=5 and 0.738 for K=10, both surpassing the highest performance recorded in the standard split. These results indicate that the model demonstrates an above-average generalization ability, effectively distinguishing between hateful and non-hateful memes. The loss curves for this K-Fold evaluation method are shown in Figures \ref{fig:kfoldtrainloss} and \ref{fig:kfoldevalloss}.

\begin{table}[t!]
  
  \caption{Model analysis recorded from 3 visual-language models as the base model}
  \label{tab:modelanalysis}
  \centering
  \begin{tabular}{cllllll}
    \toprule
    Model     & Loss & Precision & Recall & F1-micro & F1-macro & F1-weighted \\
    \midrule
    
    VisualBert [\citenum{li2019visualbert}] & 0.681 & 0.845 & 0.585 & 0.585 & 0.482 &  0.482  \\
    CLIP ViT B/32 [\citenum{radford2021learning}] & 0.682 & 0.882 & 0.574 & 0.574 & 0.496 &  0.496 \\ \midrule
    \textbf{ViLT B/32} & \textbf{0.621} & \textbf{0.872} & \textbf{0.617} & \textbf{0.617} & \textbf{0.511} & \textbf{0.581} \\
    \bottomrule
  \end{tabular}
\end{table}

\section{Discussion}
\label{discussion}
\vspace{-0.3cm}
The primary limitation of this study is the dataset size, which may limit the model's scope of learning. Expanding the dataset would enhance the model's ability to generalize across diverse contexts. Additionally, our study uses binary classification for labeling; however, adding categories such as misinformation, covert hate, and overt hate could improve the analysis's depth and accuracy. Incorporating additional modalities, such as text, audio, or video features from meme-based content, along with a combined analysis of captions and content, could provide a more comprehensive understanding of Islamophobic content.

\section{Conclusion}
\label{conclusion}
\vspace{-0.3cm}
In conclusion, this study presents a targeted approach for detecting anti-Muslim hate speech in memes using our own custom dataset and the vision-language transformer (ViLT) model. The model achieved a strong 0.738 $F_{1}$-weighted score via 10-fold cross-validation, demonstrating effective generalization, though a standard split yielded a moderate 0.581 $F_{1}$-weighted score due to overfitting. This highlights the challenges posed by subtle and complex visual hate content. Future work should expand the dataset and explore additional modalities to enhance capabilities, advancing detection strategies for more robust content moderation in digital platforms.

\bibliographystyle{plainnat}
\setcitestyle{square,numbers,comma}
\bibliography{refs.bib}






\appendix

\section{Appendix / supplemental material}



\begin{figure}[h!]
    \centering
    \includegraphics[width=0.7\textwidth]{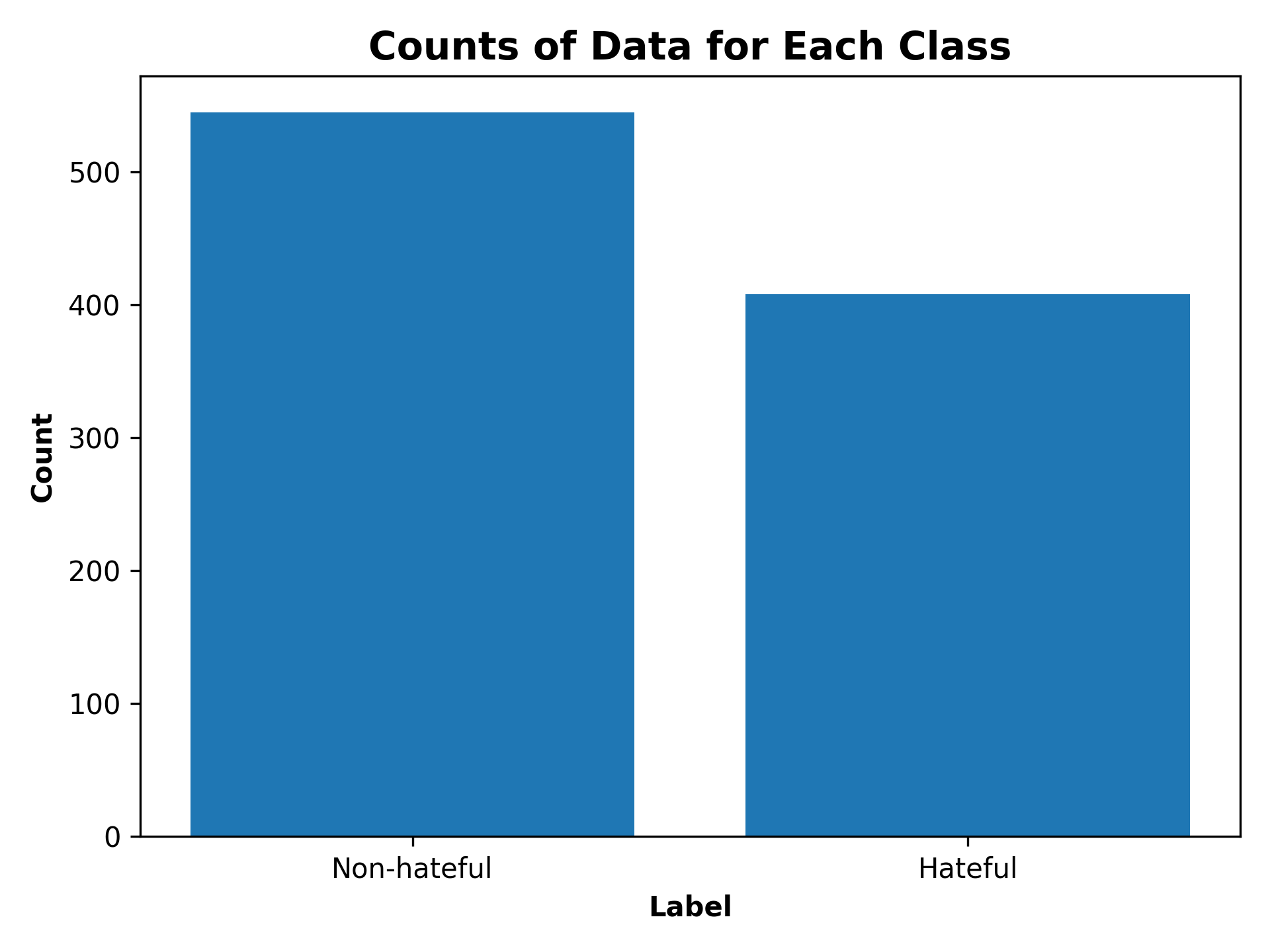}
    \caption{The distribution analysis of the classes (hateful: 1 and non-hateful: 0) in the dataset}
    \label{fig:dataset_analysis}
\end{figure}

\begin{figure}[h!]
    \centering
    \includegraphics[width=0.6\textwidth]{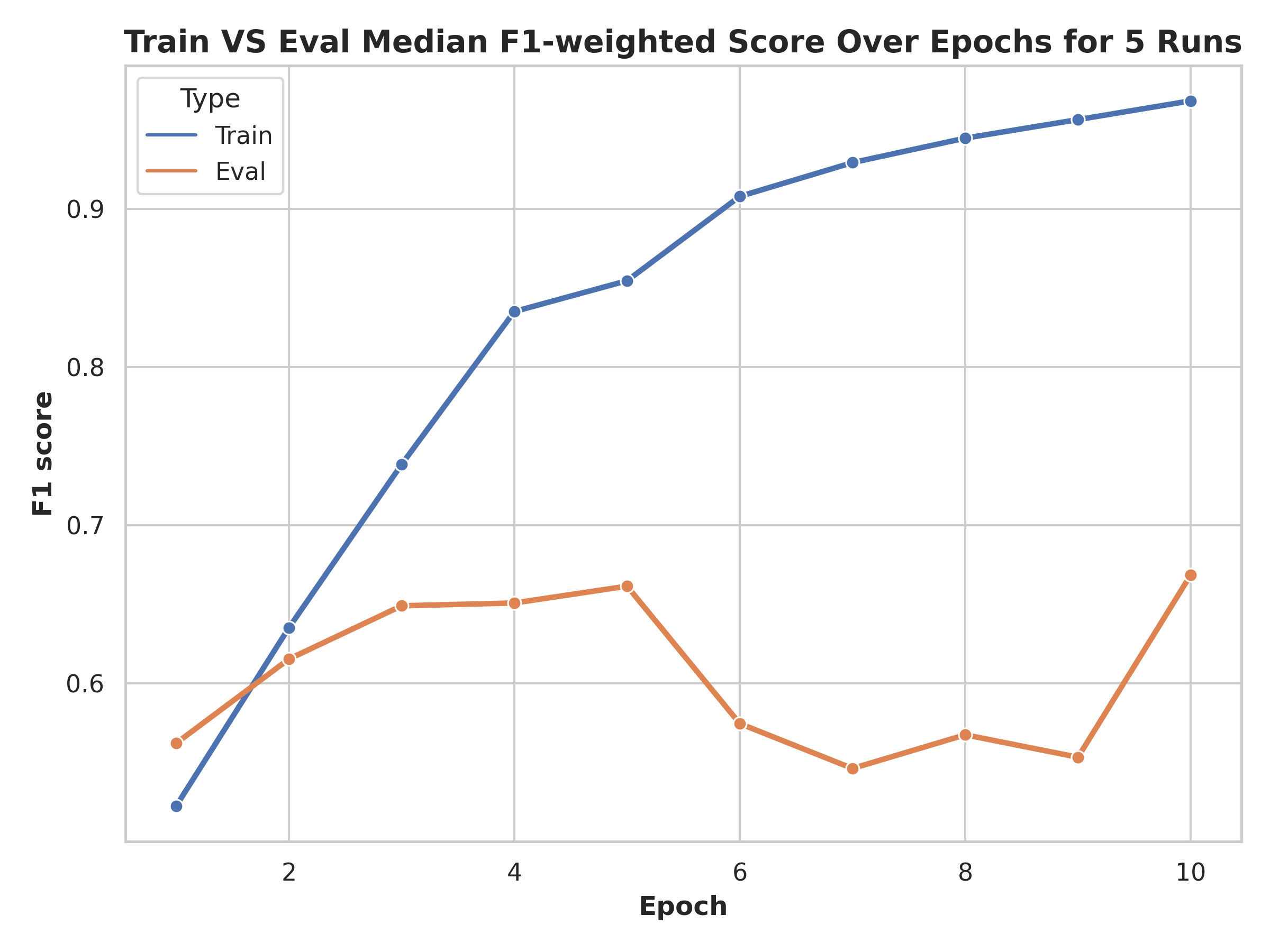}
    \caption{Median F1-weighted score curve}
    \label{fig:f1weighted}
\end{figure}

\begin{figure}[h!]
    \centering
    \includegraphics[width=0.6\textwidth]{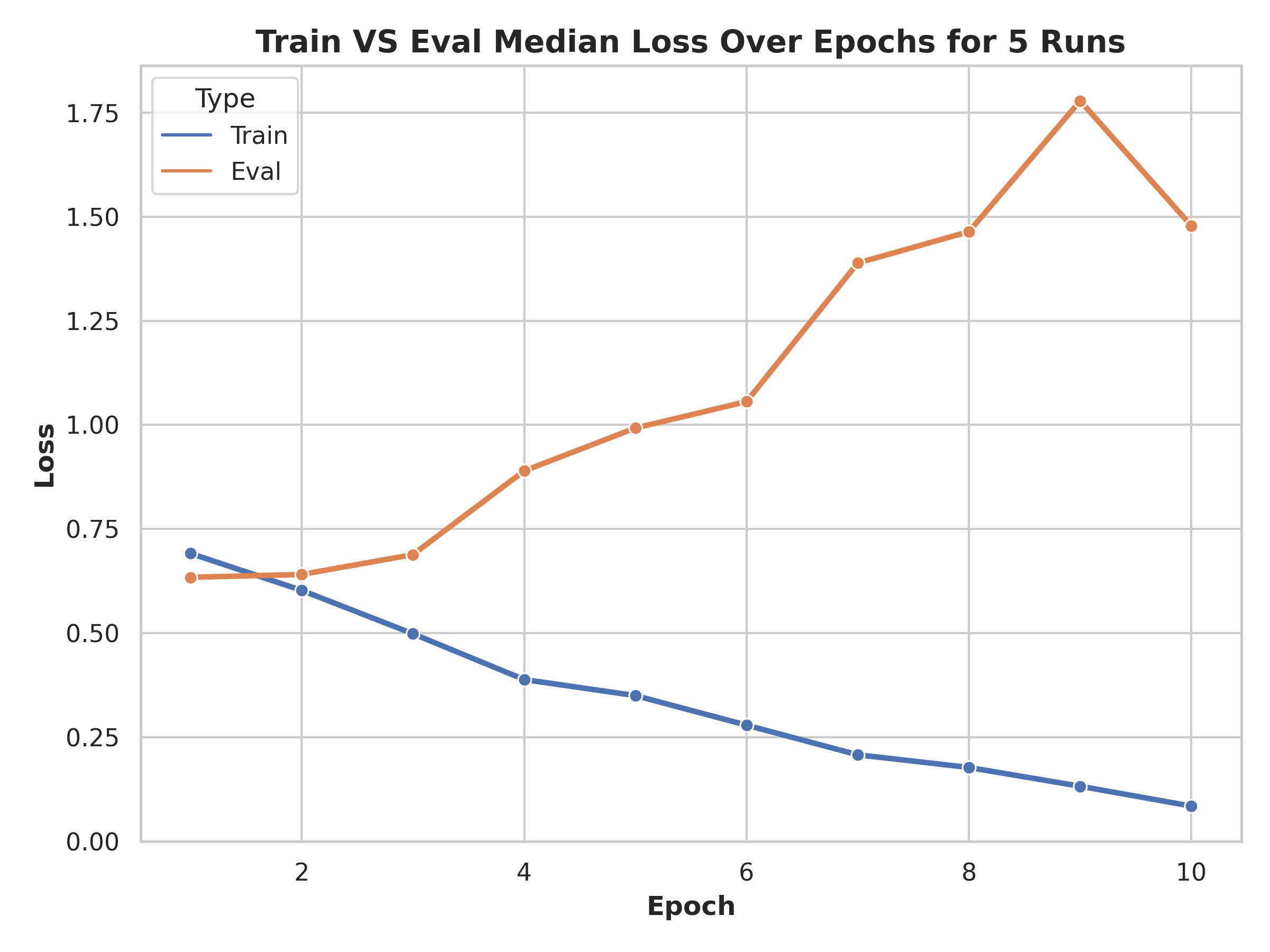}
    \caption{Median loss curve}
    \label{fig:losscurves}
\end{figure}

\begin{figure}[h!]
    \centering
    \includegraphics[width=0.6\textwidth]{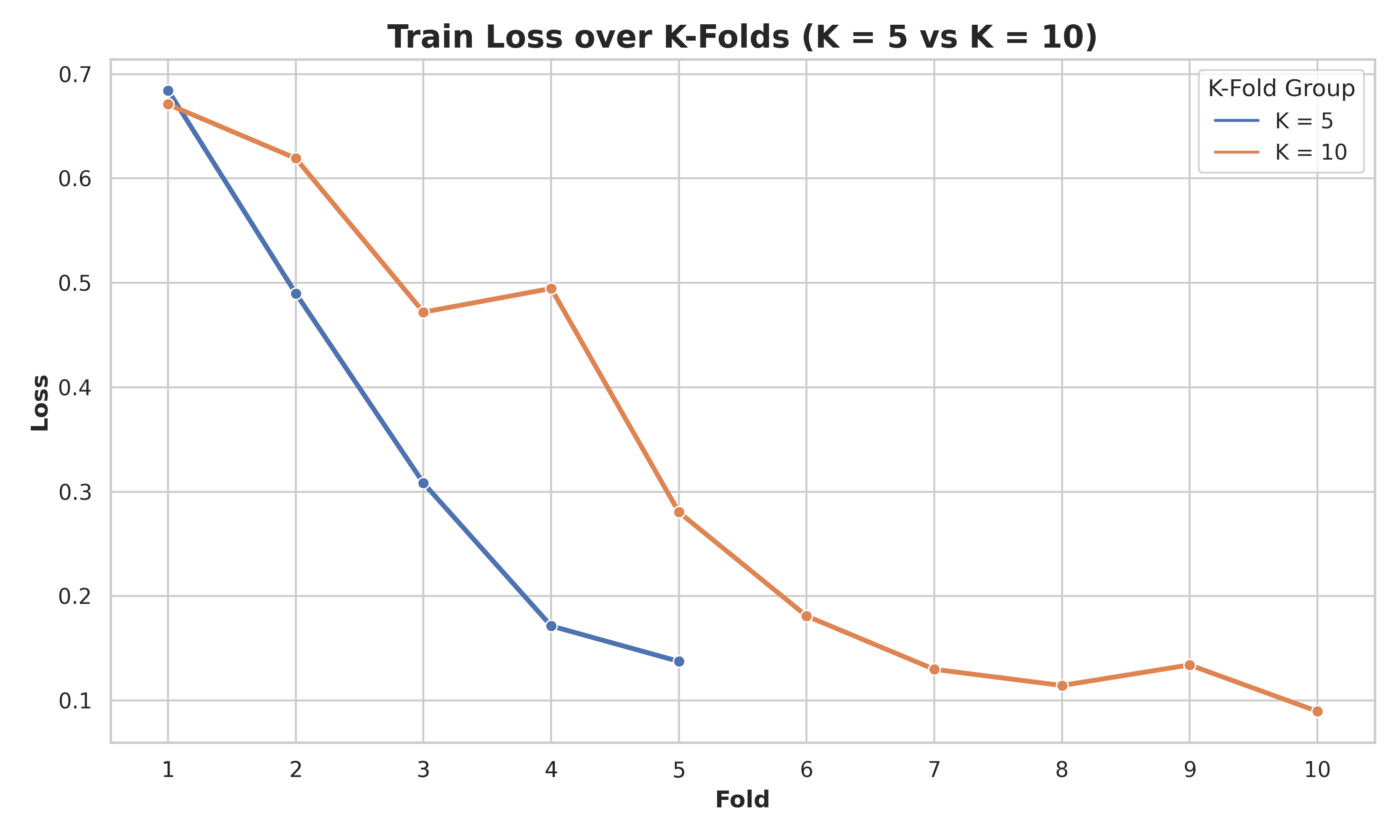}
    \caption{K-Fold train loss curves}
    \label{fig:kfoldtrainloss}
\end{figure}

\begin{figure}[h!]
    \centering
    \includegraphics[width=0.6\textwidth]{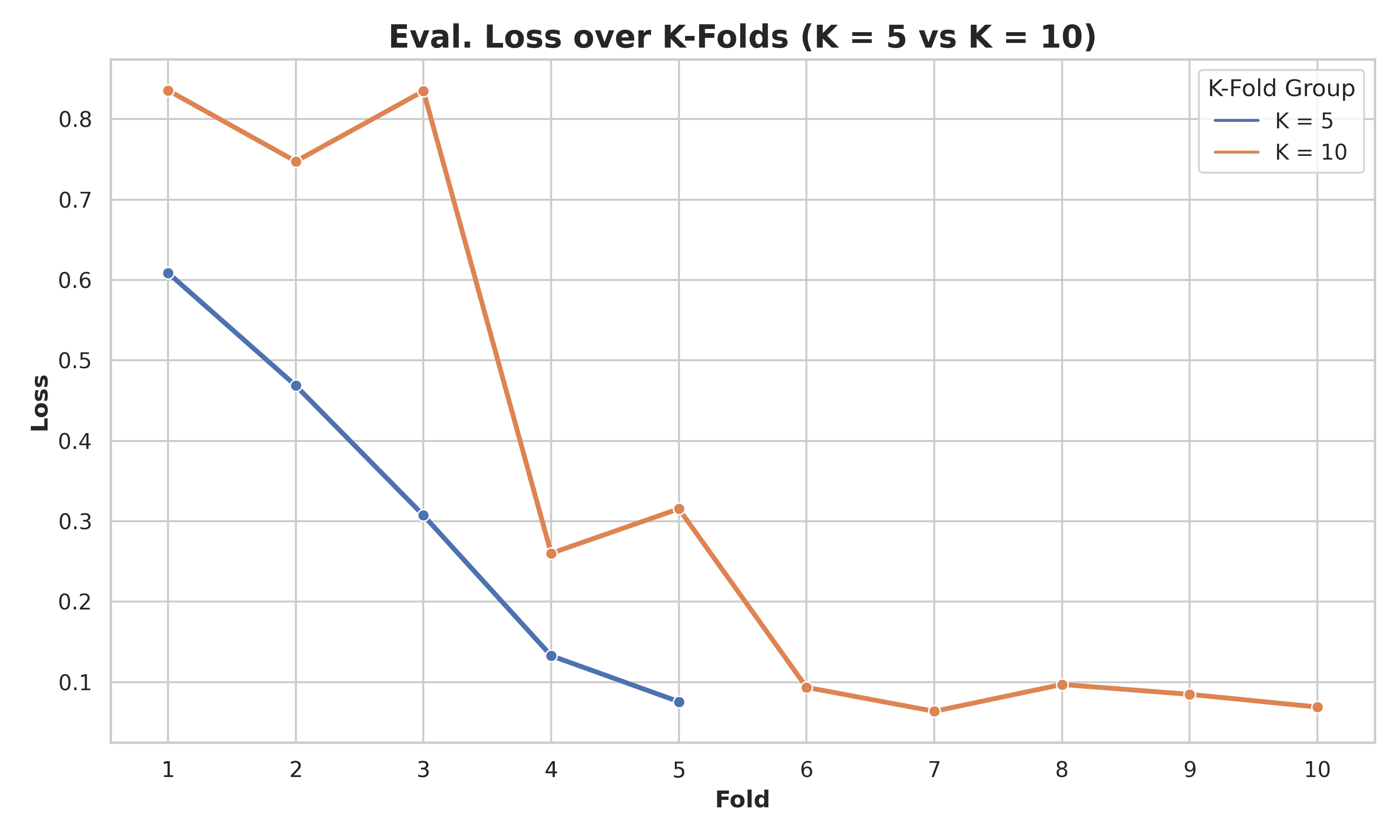}
    \caption{K-Fold validation loss curves}
    \label{fig:kfoldevalloss}
\end{figure}

\begin{table}[t!]
  
  \caption{Scores from the test set recorded from 5 runs using the train:validation:test split}
  \label{tab:traintesttable}
  \centering
  \begin{tabular}{cllllll}
    \toprule
    Run No.     & Loss & Precision & Recall & F1-micro & F1-macro & F1-weighted \\
    \midrule
    1     & 0.625 & 0.878 & 0.691 & 0.691 & 0.603 &  0.645 \\
    2     & 0.623 & 0.872 & 0.574 & 0.574 & 0.489 &  0.489  \\
    3     & 0.598 & 0.846 & 0.617 & 0.617 & 0.511 &  0.574 \\
    4     & 0.621 & 0.910 & 0.606 & 0.606 & 0.511 & 0.581 \\
    5     & 0.619 & 0.867 & 0.649 & 0.649 & 0.574 & 0.631  \\ \midrule
    Median & 0.621 & 0.872 & 0.617 & 0.617 & 0.511 & 0.581 \\
    \bottomrule
  \end{tabular}
\end{table}

\begin{table}[t!]
  \caption{Median Scores from the test set recorded for augmentation vs non-augmentation for the baseline ViLT model via train:validation:test split}
  \label{tab:augvsnonaug}
  \centering
  \begin{tabular}{ccllllll}
    \toprule
    Technique  & Loss & Precision & Recall & F1-micro & F1-macro & F1-weighted \\
    \midrule
    Non-augmentation & 0.621 & 0.872 & 0.617 & 0.617 & 0.511 & 0.581 \\
    Augmentation & 0.543 & 0.941 & 0.702 & 0.702 & 0.645 & 0.709 \\
    \bottomrule
  \end{tabular}
\end{table}

\begin{table}[t!]
  \caption{Scores from the holdout test set recorded from k-fold cross validation}
  \label{tab:kfoldtable}
  \centering
  \begin{tabular}{cllllll}
    \toprule
    K     & Loss & Precision & Recall & F1-micro & F1-macro & F1-weighted \\
    \midrule
    5     & 0.698 & 0.909 & 0.723 & 0.723 & 0.666 & 0.716 \\
    10     & 0.691 & 0.899 & 0.755 & 0.755 & 0.695 & 0.738  \\
    \bottomrule
  \end{tabular}
\end{table}

\newpage

\begin{longtable}{|>{\centering\arraybackslash}m{3cm}|p{6cm}|c|}
    \caption{Overview of Selected Dataset Samples}
    \label{tab:image_table} \\
    \hline
    \textbf{Image} & \textbf{Text} & \textbf{Label} \\
    \hline
    \endfirsthead

    \endfoot

    \includegraphics[width=2.5cm]{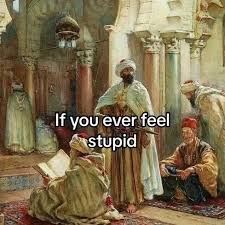} & If you ever feel stupid & 1 \\
    \hline

    \includegraphics[width=2.5cm]{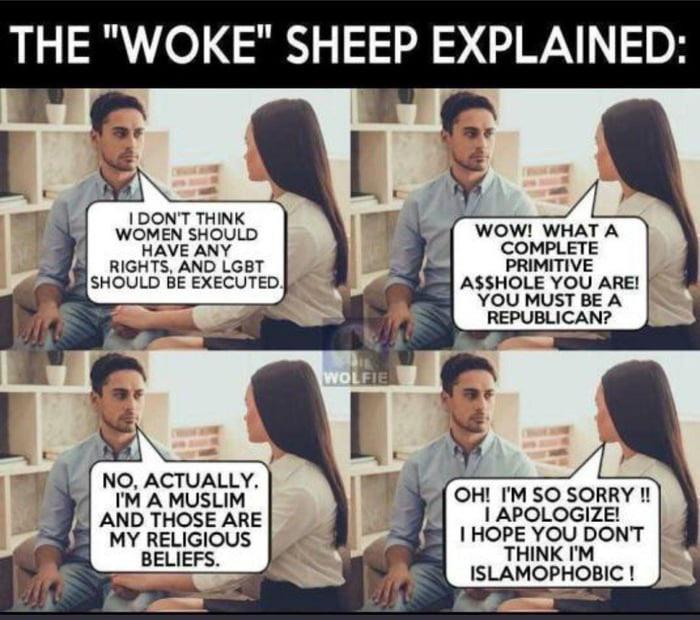} & THE "WOKE" SHEEP EXPLAINED: I DON'T THINK WOMEN SHOULD HAVE ANY RIGHTS, AND LGBTQ SHOULD BE EXECUTED. WOW! WHAT A COMPLETE PRIMITIVE A\$\$HOLE YOU ARE! YOU MUST BE A REPUBLICAN? NO, ACTUALLY I'M A MUSLIM AND THOSE ARE RELIGIOUS BELIEFS. OH! I'M SO SORRY!! I APOLOGIZE! I HOPE YOU DON'T THINK I'M ISLAMOPHOBIC!
    & 1 \\
    \hline

    \includegraphics[width=2.5cm]{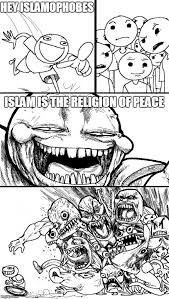} & HEY ISLAMOPHOBES ISLAM IS THE RELIGION OF PEACE & 1 \\
    \hline

    \includegraphics[width=2.5cm]{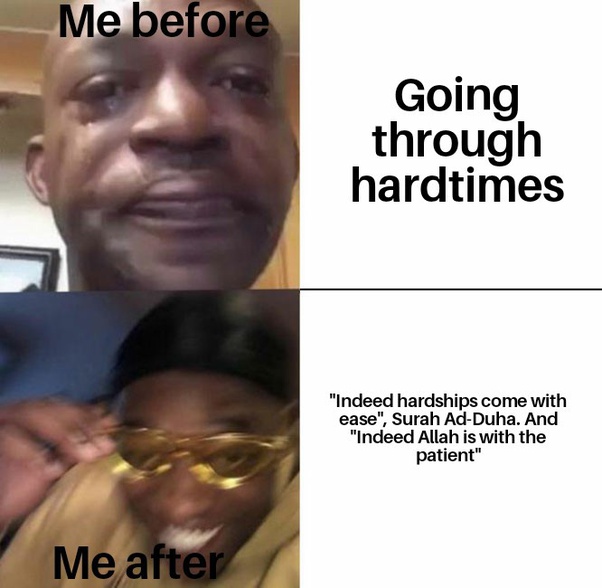} & Me before Going through hardtimes Me after "Indeed hardships come with ease", Surah Ad-Duha. And "Indeed Allah is with the patient & 0 \\
    \hline

    \includegraphics[width=2.5cm]{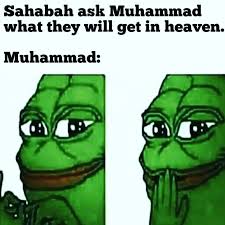} & Sahabah ask Muhammad what they will get in heaven. Muhammad: & 1 \\
    \hline
    
\end{longtable}

\end{document}